\pdfoutput=1
\documentclass{article} 

\usepackage{booktabs,csquotes,xspace,algorithm,algorithmic,enumitem,mathtools,amsmath,bm,amsmath,amsfonts,hyperref}

\newcommand{\T}[1]{\ensuremath{\mathcal{#1}}} 
\newcommand{\M}[1]{\ensuremath{\bm{#1}}} 
\newcommand{\V}[1]{\ensuremath{\bm{#1}}} 

\newcommand{\mname}{\texttt{SUSTain}\xspace}

\usepackage[binary-units=true]{siunitx} 
\begin{document}
\title{\mname: Scalable Unsupervised Scoring for Tensors and its Application to Phenotyping}

\author{Ioakeim Perros$^1$, Evangelos E. Papalexakis$^2$, Haesun Park$^1$, \\Richard Vuduc$^1$, Xiaowei Yan$^3$, Christopher deFilippi$^4$, \\Walter F. Stewart$^3$, Jimeng Sun$^1$\\Georgia Tech$^1$, UC Riverside$^2$, Sutter Health$^3$,\\ Inova Heart and Vascular Institute$^4$}

%
%
%

\maketitle

\begin{abstract}
  This paper presents a new method, which we call \mname, that extends real-valued matrix and tensor factorizations to data where values are integers.
  Such data are common when the values correspond to event counts or ordinal measures.
  The conventional approach is to treat integer data as real, and then apply real-valued factorizations.
  However, doing so fails to preserve important characteristics of the original data, thereby making it hard to interpret the results.
  Instead, our approach extracts factor values from integer datasets as \textit{scores} that are constrained to take values from a small integer set.
  These scores are easy to interpret:
  a score of zero indicates no feature contribution and higher scores indicate \textit{distinct levels} of feature importance.
  
  At its core, \mname relies on:
  a) a problem partitioning into integer-constrained subproblems, so that they can be optimally solved in an efficient manner; and
  b) organizing the order of the subproblems' solution, to promote reuse of shared intermediate results.
  We propose two variants, $\mname_M$ and $\mname_T$, to handle both matrix and tensor inputs, respectively.
We evaluate \mname against several state-of-the-art baselines on both synthetic and real Electronic Health Record (EHR) datasets. 
Comparing to those baselines, \mname shows either significantly better fit or orders of magnitude speedups that achieve a comparable fit (up to $425\times$ faster). 
We apply \mname to EHR datasets to extract patient phenotypes (i.e., clinically meaningful patient clusters).
Furthermore, $87\%$ of them were validated as clinically meaningful phenotypes related to heart failure by a cardiologist.
\end{abstract}




%


\section{Introduction}\label{sec:intro}
Matrix and tensor factorization are among 
the most promising approaches to extracting meaningful latent structure from multi-aspect data.
They have been applied successfully in diverse applications, including social network analysis~\cite{kolda2008scalable}, 
image processing~\cite{Lee1999-kk} 
and healthcare analytics~\cite{Ho2014-mh}, to name a few. 
Factorization models decompose input data into real-valued representatives revealing clusters with distinct interpretable feature profiles.

However, a significant problem arises when the input data are most naturally expressed as integer values.
Examples include event counts and ordinal data~\cite{Dong2017-sy}.
In such cases, real-valued factors distort 
the original integer characteristics. For example, real values might no longer be interpretable as counts or frequencies. Also, the possible ranges and relative differences of elements in real-valued factors is arbitrary; 
this makes it hard to intuitively compare the importance of different elements. Furthermore, in many applications, practitioners are accustomed to interpreting integer-valued scores in standardized scales. 
Real-valued factors might require arbitrary thresholding or other unnatural transformations to convert into such scales, thereby inhibiting interpretation by domain experts.

A specific motivating application for our methods is clinical phenotyping from Electronic Health Records (EHR) data.
Consider that a disease, such as heart failure, is often heterogeneous in that patients differ by underlying pathophysiology and needs.
That is, a disease is often comprised of distinct disease subtypes, or phenotypes, which vary by the ensemble of causes, associations with other diseases, and treatment needs. 
Phenotyping is intended to distinguish the latent structure among features that can, in turn, be used to prevent disease subtypes and improve treatment development and management~\cite{Richesson2016-ar}. EHR data offer a diverse and rich set of features (e.g., diagnostic, drug and procedure codes) that can serve to improve disease phenotyping.
But, these data must often be represented in integer form (e.g., clinical event counts) to be utilized in unsupervised learning. 
For example, we can construct a patient-disease matrix where the $ij$-th element represents the number of times  patient $i$ had disease $j$ documented in her records. Similarly, we can build higher-order tensors such as a patient-disease-medication one. The goal of unsupervised phenotyping is to identify patient clusters defined by unique feature sets, each one of which aligns with a distinct and intuitive clinical profile; 
in this work, we tackle this challenge via a scalable constrained integer tensor factorization.

Factorization methods have been successfully used for EHR-based unsupervised phenotyping%
~\cite{Ho2014-ml,Ho2014-mh,Wang2015-pa,Perros2015-te,Joshi2016-kt,Perros2017-dh}.
In many of those settings, the problem can be formulated via Nonnegative Matrix Factorization (NMF)~\cite{Lee1999-kk} e.g., minimizing the squared Frobenius norm of the error:
{\small
\begin{equation}\label{eq:nmf}
\text{min}\left\{ || \M{X} - \M{U} \M{V}^T ||_F^2 \;\middle|\; \M{U} \geq 0, \M{V} \geq 0 \right\}
\end{equation}}$\M{X} \in \mathbb{Z}_{+}^{M \times N}$ is a non-negative integer input matrix whose $\M{X}(i, j)$ cell reflects the event counts for the $i$-th (out of $M$) patient with respect to the $j$-th (out of $N$) features.
Given an input number $R$ of desired phenotypes, the matrix $\M{U} \in \mathbb{R}^{M \times R}$ corresponds to a membership matrix of the patients with respect to the $R$ phenotypes. And the matrix $\M{V} \in \mathbb{R}^{N \times R}$ provides the phenotypes' definition: the non-zero elements of the $r$-th column $\M{V}(:, r)$ reveal the potentially relevant features to the $r$-th phenotype. 

Interpreting those factors is crucial in order to determine whether and to what extent a patient exhibits a phenotype, as well as which set of candidate features should be considered to compose each $r$-th phenotype so that it is clinically meaningful. However, this can be challenging if the resulting factors contain arbitrary (nonnegative) real values. Real-valued factors distort the count nature of input data; thus, identifying cases and controls based on counts of relevant medical features~\cite{denny2010phewas} is no longer possible. 
Also, the possible ranges and relative differences of elements in real-valued factors is arbitrary, thus impeding the practitioner's assessment of their relative importance. 
In practice, ad hoc heuristics have been introduced with limited success: a) hard thresholding to the ranked list of factor elements, which is usually arbitrary and leads to poor model fit; b) the factor values are hidden altogether and only the elements' ranking is preserved, which  omits valuable information regarding the individual elements' actual importance. 

\begin{figure}
\centering
\includegraphics[scale=0.3]{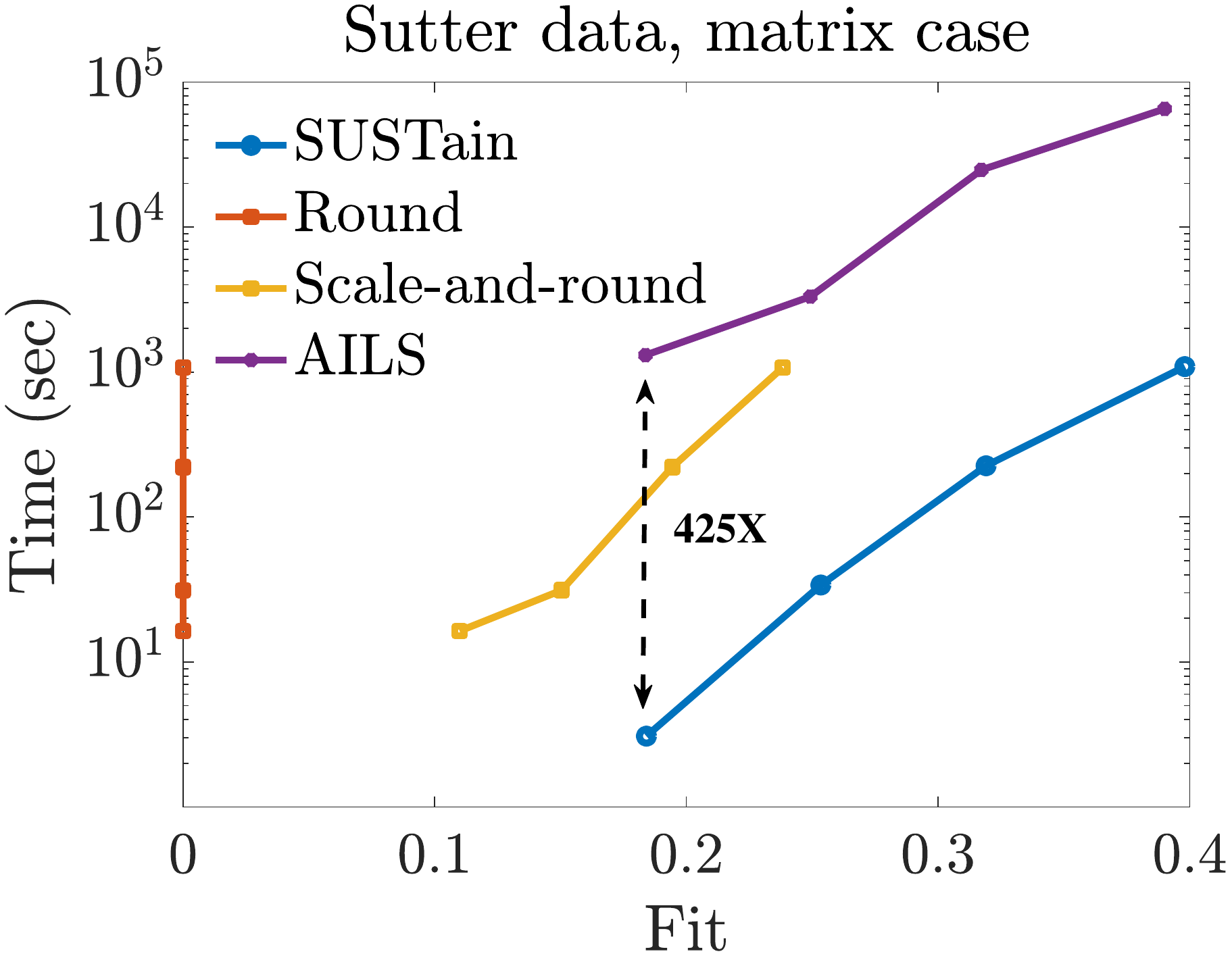}
\caption{\footnotesize Fit (range $[0, 1]$) vs time trade-off for varying target number of phenotypes $R=\{5, 10, 20, 40\}$, on a patients-by-diagnoses matrix formed of $\approx260$K patients from Sutter Palo Alto Medical Foundation Clinics. $\mname_M$ is as accurate as the most accurate baseline (based on~\cite{W_Chang2008-gt,Breen2012-ld,Dong2017-sy}), but up to $425\times$ faster ($R=5$: $\approx3$ seconds by $\mname_M$ vs. $\approx22$ minutes by AILS). Even for a larger target rank (e.g., $R=20$), $\mname_M$ is $110\times$ faster ($\approx4$ minutes by $\mname_M$ vs. $\approx7$ hours by AILS). As compared to a carefully-designed heuristic that performs a scale-and-rounding of the real-valued solution, $\mname_M$ achieves up to $16\%$ higher fit. In summary, $\mname_M$ dominates all other baselines in both time and fit. 
}
\label{fig:sustain_m_sutter_tradeoff}
\end{figure}

\noindent \textbf{Contributions:} 
To tackle these challenges, we propose Scalable Unsupervised Scoring for Tensors (\mname), a framework extracting the factor values as \textit{scores}, constrained to a small integer set. \mname offers a straightforward interpretation protocol: a score of zero indicates no feature contribution and higher scores indicate \textit{distinct levels} of feature importance.

Our methodology relies on identifying a problem partitioning into integer-constrained sub-problems so that each one of them can be solved optimally in an efficient manner; at the same time, their solution order is organized so as to promote re-use of shared intermediate results. \mname can handle both matrix and tensor inputs, through $\mname_M$ and $\mname_T$ methods, which we formulate in Sections~\ref{sec:meth_sustain_m} and \ref{sec:meth_sustain_t} respectively.

\mname yields faster and more scalable approaches than baselines achieving comparable fit, as evaluated on both synthetic (publicly-available) and real healthcare datasets. For example, as shown in Figure~\ref{fig:sustain_m_sutter_tradeoff}, $\mname_M$ achieves the same level of accuracy as the most accurate baseline \textbf{up to} $\mathbf{425\times}$ \textbf{faster}. $\mname_T$ can handle large-scale tensor inputs for which the most accurate baseline fails and scales linearly with the number of patients.

\mname's interpretation protocol is particularly meaningful for unsupervised phenotyping: it is easily understood by medical experts who are used to simple and concise, scoring-based descriptions of a patient's clinical status (e.g., risk scores\footnote{In \href{www.mdcalc.com}{MDCalc}, one can find a vast amount of such scores used in medicine.}). 
While recent work derives risk scores for predictive modeling (supervised learning)~\cite{Ustun2017-ri}, our application of \mname extracts scores for unsupervised phenotyping based on unlabeled EHR data. In Table~\ref{table:anchor_pheno}, we provide a representative phenotype extracted through our method, as part of a case study we performed on phenotyping heart failure patients. The meaningfulness of the phenotype candidates extracted through this case study was confirmed by a cardiologist, who annotated $87\%$ of them as clinically meaningful phenotypes related to heart failure. We summarize our contributions as:
\begin{itemize}[leftmargin=*]
\item \textbf{Scalable unsupervised scoring:} We propose \mname, a fast and scalable approach decomposing integer multi-aspect data into integer scores, preserving the original integer characteristics.
\item \textbf{\mname can handle matrix and tensor input:} We present \mname for both matrix (Section~\ref{sec:meth_sustain_m}) and tensor (Section~\ref{sec:meth_sustain_t}) inputs, through $\mname_M$ and $\mname_T$ methods, respectively.
\item \textbf{Evaluation on various datasets:}  We evaluate both the matrix and tensor versions on both synthetic (publicly-available) and real healthcare datasets.
\item \textbf{Phenotyping heart failure patients:} 
The interpretability of the extracted scoring-based phenotypes was confirmed by a cardiologist, who annotated $87\%$ of them as clinically meaningful.
\end{itemize}

\begin{table}
\centering
\small
\begin{tabular}{c c}
\toprule 
\textbf{Hyperlipidemia}
& \textbf{Score} \\ \midrule 
Rx\_HMG CoA Reductase Inhibitors & $3$ \\
Dx\_Disorders of lipid metabolism & $1$ \\ \bottomrule
\end{tabular}
\caption{\footnotesize Most prevalent phenotype ($26\%$ of patients) extracted via $\mname_T$ for a heart failure cohort. The $r$-th phenotype prevalence is measured through the patient membership vectors containing non-zero element in the $r$-th coordinate. The score of each feature indicates its relative frequency. The prefix for each feature indicates whether it corresponds to a medication (Rx) or a diagnosis (Dx). The cardiologist labeled the result as \enquote{hyperlipidemia} and confirmed that the two features are clinically connected to heart failure.}
\label{table:anchor_pheno}
\end{table}

\section{Background}\label{sec:back}
\begin{table}[b!]
\footnotesize
\centering
\begin{tabular}{c c} 
 \toprule 
 \textbf{Symbol} & \textbf{Definition} \\ \midrule
$\T{X}, \M{X}, \V{x}, x$ & Tensor, matrix, vector, scalar \\
$vec(\M{X})$ & Vectorization operator for matrix $\M{X}$ \\
$\Pi_C(\V{x})$ & Euclidean projection of $\V{x}$ to a set $C$ \\
$\M{X}(:, i)$ & Spans the entire $i$-th column of $\M{X}$ 
\\
$diag(\V{x})$ & Diagonal matrix with vector $\V{x}$ on the diagonal \\
$\M{X}_{(n)}$ & mode-$n$ matricization of tensor $\T{X}$ \\
$\M{A}^{(n)}$ & factor matrix corresponding to mode $n$ \\
$\circ$ & Outer product \\
$\otimes$ & Kronecker product \\
$\odot$ & Khatri-Rao product \\
$\M{A}_{\odot}^{(-n)}$ & Khatri-Rao product of all the factor matrices expect $\M{A}^{(n)}$ \\
$\M{M}^{(n)}$ & the MTTKRP corresponding to mode-$n$ \\
$*$ & Hadamard (element-wise) product \\
\bottomrule
\end{tabular}
\caption{\small Notations used throughout the paper.}
\label{tab:notations}
\end{table}
In Table~\ref{tab:notations} we summarize the notations used throughout the paper. Let $\V{x_0} \in \mathbb{R}^n$. The \textit{euclidean projection} of $\V{x_0}$ to a set $C \subseteq \mathbb{R}^n$ is defined as $\Pi_C(\V{x_0}) = \text{argmin}\left\{|| \V{x} - \V{x_0}||_2^2 \;\middle|\; \V{x}\in C \right\}$; thus, it is the problem of determining the vector $\V{x^*}$ among all $\V{x} \in C$ which is the closest to $\V{x_0}$ w.r.t. the Euclidean distance~\cite{boyd2004convex}. 
A matrix $\M{X}$ is called \textit{rank-$1$} if it can be expressed as the outer product of $2$ non-zero vectors: $\M{X} = \V{x} \circ \V{y}$. The Khatri-Rao Product (KRP) 
is the \enquote{matching column-wise} Kronecker product: for two matrices $\M{U}\in\mathbb{R}^{M\times R}, \M{V}\in\mathbb{R}^{N\times R}$ their KRP is as follows: {\small ${\M{U} \odot \M{V}} = [{\M{U}(:, 1) \otimes \M{V}(:, 1)} ~~~ {\M{U}(:, 2) \otimes \M{V}(:, 2)} ~~~ \dots ~~~ {\M{U}(:, R) \otimes \M{V}(:, R)}] \in \mathbb{R}^{M N \times R}$}

A \textit{tensor} is a multi-dimensional array. The tensor's \textit{order} denotes the number of its dimensions, also known as ways or modes (e.g.,~matrices are $2$-order tensors). A $d$-order tensor $\T{X}$ is called \textit{rank-$1$} if it can be expressed as the outer product of $d$ non-zero vectors: $\M{X} = \V{a}_1 \circ \V{a}_2 \circ \dots \circ \V{a}_d $. A \textit{fiber} is a vector extracted from a tensor by fixing all modes but one. For example, a matrix column is a mode-$1$ fiber. 
A \textit{slice} is a matrix extracted from a tensor by fixing all modes but two. 
\textit{Matricization}, also called \emph{reshaping} or \emph{unfolding}, logically reorganizes tensors into other forms without changing the values themselves. The mode-$n$ matricization of a $d$-order tensor $\T{X} \in \mathbb{R}^{I_1 \times I_2 \times \dots \times I_d}$ is denoted by $\M{X}_{(n)} \in \mathbb{R}^{I_n \times I_1 I_2 \dots I_{n-1} I_{n+1} \dots I_d}$ and arranges the mode-$n$ fibers of the tensor as columns of the resulting matrix. The Matricized-Tensor Times Khatri-Rao Product (MTTKRP)~\cite{bader2007efficient} w.r.t. mode-$n$ is the matrix multiplication $\M{X}_{(n)}~\M{A}_{\odot}^{(-n)}$, where $\M{A}_{\odot}^{(-n)}$ corresponds to the Khatri-Rao product of all the modes except the $n$-th. MTTKRP is the bottleneck operation in many sparse tensor algorithms.

\section{The \mname framework}\label{sec:meth}
First we present \mname for matrix input. Then, we describe how \mname can be extended for general high-order tensor input. Finally, we provide our interpretation protocol of \mname for unsupervised phenotyping.

\subsection{\mname for matrix input}\label{sec:meth_sustain_m}

\noindent{\bf Model:}
For an integer input matrix $\M{X}\in \mathbb{Z}_{+}^{M\times N}$ and a certain target rank $R$, the problem  can be defined as:
{\small
\begin{equation}\label{eq:sustain_m}
{\text{min}} \left\{ || \M{X} - \M{U}~\M{\Lambda}~\M{V}^T ||_F^2 \;\middle|\;
\M{U}\in \mathbb{Z}_{\tau}^{M\times R},\M{V} \in \mathbb{Z}_{\tau}^{N\times R}, \M{\Lambda}\in \mathbb{Z}_{+}^{R\times R} \right\}
\end{equation}}where $\mathbb{Z}_{\tau} = \{0, 1, \dots, \tau\}$ is the set of nonnegative integers up to $\tau$, 
$\mathbb{Z}_{+} = \{ 1, 2, \dots, \infty\}$ is the set of positive integers and $\M{\Lambda}$ is a diagonal $R$-by-$R$ matrix. The above problem can be also formulated as $|| \M{X} - \sum_{r=1}^R \V{\lambda}(r)~\M{U}(:, r)~\M{V}(:, r)^T ||_F^2$ where $\V{\lambda}(r)=\M{\Lambda}(r, r)$. The reason for having $\V{\lambda}(r)$ is to absorb any scaling of each $r$-th rank-1 component, since the entries of $\M{U}$ and $\M{V}$ factors are upper bounded by $\tau$. Note that the $\V{\lambda}(r)$ values cannot be simply obtained through normalization as in the corresponding real-valued models (e.g., NMF~\cite{Lee1999-kk}), due to the integer constraints.
Finally, note that the integer set $\mathbb{Z}_{\tau}$ can easily vary for different factor matrices and even allow negative integers; this can also happen for the input matrix $\M{X}$. The formulation in Problem~\eqref{eq:sustain_m} favors simplicity of presentation and matches the need of phenotyping applications.

\noindent{\bf Fitting Algorithm:}
We employ an alternating updating 
scheme to tackle the non-convex optimization Problem~\eqref{eq:sustain_m}. 
Our scheme leads to optimal solutions to each one of the sub-problems in an efficient manner, while organizing the order of updates so as to promote re-use of already computed intermediate results.

We follow the intuition behind the Hierarchical Alternating Least Squares (HALS) framework, which enables isolating and solving for each $k$-th rank-$1$ component separately. 
Thus, 
Problem~\eqref{eq:sustain_m} gives:
{\small
\begin{align}\label{eq:sustain_m_r1}
{\text{min}}& \{|| \underbrace{\M{X} - \sum_{r=1, r\neq k}^R \V{\lambda}(r)~\M{U}(:, r)~\M{V}(:, r)^T}_{\M{R}_k} - \V{\lambda}(k)~\M{U}(:, k)~\M{V}(:, k)^T ||_F^2 \\
& \; | \; \M{U}\in \mathbb{Z}_{\tau}^{M\times R},\M{V} \in \mathbb{Z}_{\tau}^{N\times R}, \M{\Lambda}\in \mathbb{Z}_{+}^{R\times R} \nonumber \}
\end{align}}where $\M{R}_k$ corresponds to the \enquote{residual matrix} and is considered fixed when solving for the $k$-th rank-1 component. The objective can be written as~\cite{Kolda1998-up}:
{\small
\begin{equation*}
J = ||\M{R}_k ||_F^2 + \V{\lambda}^2(k)~ 
~||\M{U}(:, k)||_2^2 ~ ||\M{V}(:, k)||_2^2 - 2~\V{\lambda}(k)~\M{U}(:, k)^T~\M{R}_k~\M{V}(:, k)
\end{equation*}} We set:
{\small\[ \partial J / \partial \V{\lambda}(k) = 2~{\V{\lambda}(k)}~ 
~{||\M{U}(:, k)||_2^2} ~ {||\M{V}(:, k)||_2^2} - 2~{\M{U}(:, k)^T}~{\M{R}_k}~{\M{V}(:, k)} = 0 \]} and obtain:
{\small
\[ \lambda_k^* := \frac{\M{U}(:, k)^T~\M{R}_k~\M{V}(:, k)}{||\M{U}(:, k)||_2^2 ~ ||\M{V}(:, k)||_2^2} \]
}If $\M{U}(:, k)^T~\M{R}_k~\M{V}(:, k) > 0$ then the minimum value of $J$ for $\V{\lambda}(k) \in \mathbb{Z}_{+}$ is obtained at $max(1, round(\lambda_k^*))$ where $round()$ rounds to the nearest integer. If $\M{U}(:, k)^T~\M{R}_k~\M{V}(:, k) \leq 0$, then the minimum objective value for 
$\V{\lambda}(k) \in \mathbb{Z}_{+}$ is attained at $\V{\lambda}(k) = 1$. Combining these two cases, 
the optimal $\V{\lambda}(k) \in \mathbb{Z}_{+}$ is given by:
{\small
\begin{equation}\label{eq:sustain_m_opt_lambda_1}
\V{\lambda}(k) \gets max\left(1, round\left(\frac{\M{U}(:, k)^T~\M{R}_k~\M{V}(:, k)}{||\M{U}(:, k)||_2^2 ~ ||\M{V}(:, k)||_2^2}\right)\right)
\end{equation}}In practice, $\M{R}_k$ may be large ($M \times N$) and dense, even if the input is sparse (as happens in our main motivating application); thus its explicit materialization should be avoided~\cite{gillis2011nonnegative,Kim2014-gw}. 
Expanding the above expression gives:
{\footnotesize
\begin{equation}\label{eq:sustain_m_opt_lambda_2}
\V{\lambda}(k) \gets max\left(1, round\left(  \V{\lambda}(k) +  \frac{ \M{V}(:, k)^T ~ \left( [\M{X}^T~\M{U}]_{:, k} ~-~ 
\M{V}~\M{\Lambda}~
[\M{U}^T~\M{U}]_{:,k} \right)   }
{[\M{U}^T~\M{U}]_{k,k}~[\M{V}^T~\M{V}]_{k,k}}\right)\right)
\end{equation}}
Next, solving Problem~\eqref{eq:sustain_m_r1} for $\M{V}(:, k)$ gives:
{\small
\begin{equation}\label{eq:sustain_m_v}
{\text{min}} \left\{ || \M{R}_k - \V{\lambda}(k)~\M{U}(:, k)~\M{V}(:, k)^T ||_2^2 \;\middle|\;
\M{V}(:, k) \in \mathbb{Z}_{\tau}^N \right\}
\end{equation}}To solve the above, we apply the Optimal Scaling Lemma~\cite{bro1998least} for the integer constraint. This Lemma states that for any set $C$ of constraints imposed on $\V{b}$, 
it holds that:
{\small
\begin{equation*}
{\text{min}} \left\{ || \M{Y} - \V{x}~\V{b}^T ||_2^2 \;\middle|\;
\V{b} \in C \right\} \Leftrightarrow \Pi_{C}(\V{\beta})
\end{equation*}}where $\V{\beta} = \frac{\V{x}^T~\M{Y}}{\V{x}^T~\V{x}}$ is the unconstrained solution to the above problem. This means that the optimal solution of the constrained problem is simply the projection of the unconstrained solution onto the constraint set $C$. 
Thus, the optimal solution of Problem~\eqref{eq:sustain_m_v} is:
{\small
\begin{equation}\label{eq:sustain_m_v1}
\M{V}(:, k) \gets \mathbf{\Pi}_{\mathbb{Z}_{\tau}^N}\left(\frac{\M{R}_k^T~\M{U}(:, k)}{ [\M{U}^T~\M{U}]_{k,k} ~ \lambda(k)}\right)
\end{equation}}Since $\mathbb{Z}_{\tau}^N$ is the Cartesian product of subsets of the real line, i.e., $\mathbb{Z}_{\tau}^N = \underbrace{\mathbb{Z}_{\tau} \times \mathbb{Z}_{\tau} \times \dots \times \mathbb{Z}_{\tau}}_{\text{$N$ times}}$ we can take
{\small
\begin{equation}\label{eq:sustain_m_v2}
\Pi_{\mathbb{Z}_{\tau}^N}(\M{V}(:, k)) = [ \Pi_{\mathbb{Z}_{\tau}}(\M{V}(1, k)) , \dots, 
\Pi_{\mathbb{Z}_{\tau}}(\M{V}(N, k))]
\end{equation}}thus project each scalar coordinate individually. For a real-valued scalar $\alpha$, projecting onto $\mathbb{Z}_{\tau}$ gives~\cite{takapoui2017simple}:
{\small
\begin{equation}\label{eq:sustain_m_v3}
\Pi_{\mathbb{Z}_{\tau}}(\alpha) = min\left( max\left( round\left(\alpha \right), 0 \right), \tau  \right)
\end{equation}}Finally, expanding $\M{R}_k$ in Expression~\eqref{eq:sustain_m_v1}, combining with \eqref{eq:sustain_m_v2}, \eqref{eq:sustain_m_v3} and setting:
{\small
\begin{equation}\label{eq:sustain_m_v_final_a}
\V{b} \gets \M{V}(:, k) + \frac{[\M{X}^T~\M{U}]_{:, k} - \M{V}~\M{\Lambda}~[\M{U}^T~\M{U}]_{:,k} }
{[\M{U}^T~\M{U}]_{k,k}~\V{\lambda}(k)}
\end{equation}}
gives the optimal solution for Problem~\eqref{eq:sustain_m_v}:
\begin{equation}\label{eq:sustain_m_v_final}
\M{V}(:, k) \gets min\left( max\left( round\left( 
\V{b} \right), 0 \right), \tau  \right)
\end{equation}
where $min(), max(), round()$ are taken element-wise.

Having derived the updates for $\V{\lambda}(k), \M{V}(:, k)$, in Relations~\eqref{eq:sustain_m_opt_lambda_2} and \eqref{eq:sustain_m_v_final} respectively, we remark that the computationally expensive intermediate results $[\M{X}^T~\M{U}]$ and $[\M{U}^T~\M{U}]$ are shared between them. To exploit that, we choose to successively update $\V{\lambda}(k)$ and $\M{V}(:, k)$ during the same iteration and iterate $\forall k \in \{1,\dots, R\}$. As a result of the proposed update order, the only non-negligible additional operation in order to compute both $\V{\lambda}(k)$ and $\M{V}(:, k)$ is to re-compute $\V{t} := {\M{V}~\M{\Lambda}~[\M{U}^T~\M{U}]_{:,k}}$ after having updated $\V{\lambda}(k)$. 

Re-computing $\V{t}$ can be further optimized by observing that only the contribution of the $k$-th component $\V{t_k} := {\M{V}(:, k)~*~\V{\lambda}(k)~[\M{U}^T~\M{U}]_{k,k}}$ has to be adjusted. Thus, we can store $\V{t}$ and $\V{t_k}$, compute $\V{\lambda}'(k)$, and then adjust $\V{t}$ as: $\V{t} \gets \V{t} - \V{t_k} + \left({\M{V}(:, k)~*~\V{\lambda}'(k)~[\M{U}^T~\M{U}]_{k,k}} \right)$. 

Updating $\M{U}(:, k)$ can be executed in symmetric fashion to $\M{V}(:, k)$. In Algorithm~\ref{alg:sustain_m}, we present our main procedure to update both the factor matrices $\M{U}$ and $\M{V}$ and $\V{\lambda}$ values in an alternating fashion. In Algorithm~\ref{alg:core}, we provide the definition of $\mname$\_Update\_Factor which updates a single factor (denoted as $\M{F}$) and the vector $\V{\lambda}$.
\begin{algorithm}
    \caption{$\mname_M$}
    \label{alg:sustain_m}
    \small
    \begin{algorithmic}[1]
    	\REQUIRE $\M{X} \in \mathbb{R}^{M \times N}$, target rank $R$ and upper bound $\tau$
    	\ENSURE $\M{U}\in \mathbb{Z}_{\tau}^{M\times R}, \M{V}\in \mathbb{Z}_{\tau}^{N\times R}, \V{\lambda}\in\mathbb{Z}_{+}^{R}$
        \STATE Initialize $\M{U}, \M{V}, \M{\Lambda}$
        \WHILE{convergence criterion is not met}
        	\STATE $\M{F} \gets \M{U}, \M{M} \gets \M{X}~\M{V}, \M{C} \gets \M{V}^T~\M{V}$
			\STATE $[\M{U}, \V{\lambda}] = \mname\text{\_Update\_Factor}(\M{F}, \M{M}, \M{C}, \V{\lambda}, R, \tau)$ 
            \STATE $\M{F} \gets \M{V}, \M{M} \gets \M{X}^T~\M{U}, \M{C} \gets \M{U}^T~\M{U}$
            \STATE $[\M{V}, \V{\lambda}] = \mname\text{\_Update\_Factor}(\M{F}, \M{M}, \M{C}, \V{\lambda}, R, \tau)$ 
        \ENDWHILE
    \end{algorithmic}
\end{algorithm}
\begin{algorithm}
    \caption{$\mname$\_Update\_Factor$(\M{F}, \M{M}, \M{C}, \V{\lambda}, R, \tau)$}
    \label{alg:core}
    \small
    \begin{algorithmic}[1]
    	\REQUIRE $\M{F} \in \mathbb{ \mathbb{Z}_{\tau}}^{I \times R}, \M{M} \in \mathbb{ \mathbb{R}}^{I \times R}, \M{C} \in \mathbb{ \mathbb{R}}^{R \times R}, \V{\lambda} \in  \mathbb{Z}_{+}^R $, target rank $R$ and upper bound $\tau$
    	\ENSURE $\M{F} \in \mathbb{ \mathbb{Z}_{\tau}}^{I \times R}, \V{\lambda} \in  \mathbb{Z}_{+}^R $
        	\FOR{$k=1, \dots, R$}
            	\STATE $\V{t} \gets \M{F}~
                \left(\V{\lambda}*\M{C}(:, k) \right)$
                \STATE $\V{t_k} \gets \M{F}(:, k)~*~\V{\lambda}(k)~\M{C}(k, k)$
            	\STATE $\alpha \gets \V{\lambda}(k) +
                \frac{ \M{F}(:, k)^T \left(\M{M}(:, k)~-~
   \V{t}\right) }
{\M{C}(k, k)~[\M{F}^T~\M{F}]_{k,k}}$
				\STATE $\lambda(k) \gets max\left(1, round\left(  \alpha \right)\right)$
                \STATE $\V{t} \gets \V{t} - \V{t_k} + \left( \M{F}(:, k)~*~\V{\lambda}(k)~\M{C}(k, k) \right)$
                \STATE $\V{b} \gets \M{F}(:, k) + \frac{ \M{M}(:, k)~-~\V{t} }
{C(k,k)~\V{\lambda}(k)}$
				\STATE $\M{F}(:, k) \gets min\left( max\left( round\left( 
\V{b} \right), 0 \right), \tau  \right)$
            \ENDFOR
    \end{algorithmic}
\end{algorithm}

\noindent \textbf{Computational Complexity:} 
The asymptotic cost of executing Algorithm~\ref{alg:core} is $2 R^2 I$ flops (i.e., floating-point operations), $\forall R > 5$. This step costs $2 R^2 N$ when updating $\M{V}$ and $2 R^2 M$ when updating $\M{U}$. In Algorithm~\ref{alg:sustain_m}, assuming the input $\M{X}$ is sparse, the cost of each one of $\M{X}~\M{V}$ and $\M{X}^T~\M{U}$ is $2~ nnz(\M{X})~R$ flops. 
Also, computing $\M{V}^T~\M{V}$ and $\M{U}^T~\M{U}$ cost $2 R^2 N$ and $2 R^2 M$ flops respectively. Thus, the total cost is: $4 R \left(nnz(\M{X}) + (M+N) R\right)$ flops. 

\subsection{\mname for tensor input}\label{sec:meth_sustain_t}
\noindent{\bf Model:}
For a tensor $\T{X}\in \mathbb{R}^{I_1\times I_2 \times \dots I_d}$ of order $d$ and a certain target rank $R$, the problem can be  defined as 
{\small
\begin{multline}\label{eq:sustain_t}
{\text{min}} \{ || \T{X} - 
\sum_{r=1}^R \V{\lambda}(r)~\M{A}^{(1)}(:, r)~\circ
\dots~\circ~\M{A}^{(d)}(:, r) ||_F^2 \\ | \;
\M{A}^{(n)}\in \mathbb{Z}_{\tau}^{I_n\times R}, \V{\lambda}(r) \in \mathbb{Z}_{+} \}
\end{multline}}where $n= \{1, \dots, d \}, \mathbb{Z}_{\tau} = \{0, 1, \dots, \tau\}$ is the set of nonnegative integers up to $\tau$ and $\mathbb{Z}_{+} = \{ 1, 2, \dots, \infty\}$ is the set of positive integers. Our model is an extension of $\mname_M$ presented in Section~\ref{sec:meth_sustain_m} for high-order tensors. It can be viewed as a constrained version of the CP tensor 
model~\cite{harshman1970foundations,carroll1970analysis}.

\noindent{\bf Fitting Algorithm:}
Similarly to the matrix case, we set:
{\small \[ \T{R}_k := \T{X} - 
\sum_{r=1, r\neq k}^R \V{\lambda}(r)~\M{A}^{(1)}(:, r)~\circ
\dots~\circ~\M{A}^{(d)}(:, r) \]}
Thus, Problem~\eqref{eq:sustain_t} becomes:
{\small
\begin{multline}\label{eq:sustain_t1}
{\text{min}} \{ || \T{R}_k - 
 \V{\lambda}(k)~\M{A}^{(1)}(:, k)~\circ
\dots~\circ~\M{A}^{(d)}(:, k) ||_F^2 \\ | \;
\M{A}^{(n)}\in \mathbb{Z}_{\tau}^{I_n\times R}, \V{\lambda}(k) \in \mathbb{Z}_{+} \}
\end{multline}}We matricize the above expression w.r.t. mode-$n$ and utilize the fact that the mode-$n$ matricization of a rank-$1$ tensor ${\V{b}_1 \circ \dots \circ \V{b}_d}$ can be expressed as ${\V{b}_n \left(\V{b}_d \otimes \dots \otimes \V{b}_{n+1} \otimes \V{b}_{n-1}  \otimes \dots \otimes \V{b}_1 \right)^T}$~\cite{golub2013matrix}:
{\small
\begin{multline}\label{eq:sustain_t2}
{\text{min}} \{ || {\T{R}_k}_{(n)} - 
 \V{\lambda}(k)~\M{A}^{(n)}(:, k) \\
(\M{A}^{(d)}(:, k) \otimes \dots \otimes \M{A}^{(n+1)}(:, k) \otimes \M{A}^{(n-1)}(:, k)  \otimes \dots \otimes \M{A}^{(1)}(:, k) )^T ||_F^2 \\ | \;
\M{A}^{(n)}\in \mathbb{Z}_{\tau}^{I_n\times R}, \V{\lambda}(k) \in \mathbb{Z}_{+} \}
\end{multline}}We set $\M{A}_{\odot}^{(-n)} :=  \M{A}^{(d)} \odot \dots \odot \M{A}^{(n+1)} \odot \M{A}^{(n-1)}  \odot \dots \odot \M{A}^{(1)} $ as the Khatri-Rao Product of all the factor matrices except the $n$-th and 
{\small
\begin{equation}\label{eq:sustain_t_coeff}
\M{C}^{(-n)} := {\M{A}^{(d)}}^T~\M{A}^{(d)}  * \dots * {\M{A}^{(n+1)}}^T~\M{A}^{(n+1)}  * {\M{A}^{(n-1)}}^T~\M{A}^{(n-1)} * \dots * {\M{A}^{(1)}}^T~\M{A}^{(1)}
\end{equation}}as the Hadamard product of the Gram matrices of all the factor matrices except the $n$-th. Then, Objective~\eqref{eq:sustain_t2} becomes 
{\small
\begin{equation}\label{eq:sustain_tensor}
{\text{min}} \{ || {\T{R}_k}_{(n)} - \V{\lambda}(k)~\M{A}^{(n)}(:, k)~
\M{A}_{\odot}^{(-n)}(:, k)^T ||_F^2  | \;
\M{A}^{(n)}\in \mathbb{Z}_{\tau}^{I_n\times R}, \V{\lambda}(k) \in \mathbb{Z}_{+} \}
\end{equation}} Solving the above for $\V{\lambda}(k)$ can be handled equivalently to the corresponding matrix case (Relation~\eqref{eq:sustain_m_opt_lambda_1}). Thus, the optimal solution for $\V{\lambda}(k) \in \mathbb{Z}_{+}$ is:
{\small
\begin{equation}\label{eq:sustain_t_opt_lambda_extra}
\V{\lambda}(k) \gets max\left(1, round\left(
\frac{ \M{A}^{(n)}(:, k)^T~
{\T{R}_k}_{(n)}~
\M{A}_{\odot}^{(-n)}(:, k) }
{ || \M{A}^{(n)}(:, k) ||_2^2~
||\M{A}_{\odot}^{(-n)}(:, k) ||_2^2 }
\right)\right)
\end{equation}}By exploiting that~\cite{kolda2009tensor}:
{\small
\[ ||\M{A}_{\odot}^{(-n)}(:, k) ||_2^2 = \M{A}_{\odot}^{(-n)}(:,k)^T \M{A}_{\odot}^{(-n)}(:,k) = [{\M{A}_{\odot}^{(-n)}}^T~\M{A}_{\odot}^{(-n)}]_{k,k} = \M{C}^{(-n)}(k, k) \]} and expanding:
{\small
\begin{equation}\label{eq:sustain_t4}
{\T{R}_k}_{(n)}~
\M{A}_{\odot}^{(-n)}(:, k) = \M{M}^{(n)}(:, k) - \M{A}^{(n)}~\M{\Lambda}~
\M{C}^{(-n)}(:, k) + \V{\lambda}(k)~
\M{C}^{(-n)}(k, k)~ \M{A}^{(n)}(:, k)
\end{equation}}
where $\M{M}^{(n)}(:, k)$ is the Matricized-Tensor Times Khatri-Rao Product (MTTKRP)~\cite{bader2007efficient} operation w.r.t. mode $n$, we get the optimal solution for $\V{\lambda}(k) \in \mathbb{Z}_{+}$ as:
{\footnotesize
\begin{equation}\label{eq:sustain_t_lambdak}
\V{\lambda}(k) \gets max\left(1, round\left( \V{\lambda}(k) + \frac{ \M{A}^{(n)}(:, k)^T
\left( \M{M}^{(n)}(:, k)~-~
\M{A}^{(n)}~\M{\Lambda}~\M{C}^{(-n)}(:,k) \right)}
{
\M{C}^{(-n)}(k, k)~
[{ \M{A}^{(n)}}^T  \M{A}^{(n)}]_{k,k}
} \right) \right)
\end{equation}}Next, we transpose the Objective~\eqref{eq:sustain_tensor} and solving for $\M{A}^{(n)}(:, k)$ can be handled as in the matrix case (Relation~\eqref{eq:sustain_m_v1}) through the Optimal Scaling Lemma~\cite{bro1998least}. Thus, the optimal $\M{A}^{(n)}(:, k) \in \mathbb{Z}_{\tau}^{I_n}$ is given by:
{\small
\begin{equation}\label{eq:sustain_tensor_v_solution}
\M{A}^{(n)}(:, k) \gets \mathbf{\Pi}_{\mathbb{Z}_{\tau}^{I_n}}\left(
\frac{ {\T{R}_k}_{(n)}~ \M{A}_{\odot}^{(-n)}(:, k) }
{ \M{C}^{(-n)}(k,k) ~ \lambda(k)}
\right) 
\end{equation}
}
Finally, combining Equation~\eqref{eq:sustain_t4} into the above gives:
{\small
\begin{equation}\label{eq:sustain_t_ak}
\M{A}^{(n)}(:, k) \gets \mathbf{\Pi}_{\mathbb{Z}_{\tau}^{I_n}}\left(
\M{A}^{(n)}(:, k) +
\frac{\M{M}^{(n)}(:, k) - \M{A}^{(n)}~
\M{\Lambda}~\M{C}^{(-n)}(:, k)}
{\M{C}^{(-n)}(k, k)~\V{\lambda}(k)}\right)
\end{equation}}Note the direct correspondence of the above formulations for ${\V{\lambda}(k)}$, ${\M{A}^{(n)}(:, k)}$ with the core update Algorithm~\ref{alg:core} we used for the matrix case. If we set ${\M{F} \gets \M{A}^{(n)}, \M{M} \gets \M{M}^{(n)}, \M{C} \gets \M{C}^{(-n)}}$ then we can simply use Algorithm~\ref{alg:core} to update a single factor $\M{A}^{(n)}$ and the $\V{\lambda}$ values. Also, we can exploit the development of existing scalable software libraries computing the bottleneck MTTKRP kernel for sparse data efficiently~\cite{bader2007efficient}. In Algorithm~\ref{alg:sustain_t}, we summarize the operations of our methodology for tensor input.

\begin{algorithm}
    \caption{$\mname_T$}
    \label{alg:sustain_t}
    \small
    \begin{algorithmic}[1]
    	\REQUIRE $\T{X}\in \mathbb{R}^{I_1\times I_2 \times \dots I_d}$, target rank $R$ and upper bound $\tau$
    	\ENSURE $\M{A}^{(n)}\in \mathbb{Z}_{\tau}^{I_n\times R},$ with $n\in \{1, \dots, d\}, \V{\lambda}\in\mathbb{Z}_{+}^{R}$
        \STATE Initialize $\M{A}^{(n)}, \V{\lambda}$
        \WHILE{convergence criterion is not met}
		\FOR{$n = 1, \dots, d$}  
        	\STATE $\M{M}^{(n)} \gets             \T{X}_{(n)}~\M{A}_{\odot}^{(-n)}$ \hspace{4em}//  \textit{MTTKRP}
            \STATE Compute $\M{C}^{(-n)}$ as in Relation~\eqref{eq:sustain_t_coeff}
			\STATE $[\M{A}^{(n)}, \V{\lambda}] = \mname\text{\_Update\_Factor}(\M{A}^{(n)},~\M{M}^{(n)},
            ~\M{C}^{(-n)}, \V{\lambda}, R, \tau)$
        \ENDFOR
        \ENDWHILE
    \end{algorithmic}
\end{algorithm}

\noindent \textbf{Computational Complexity:} Updating the $n$-th mode in Algorithm~\ref{alg:sustain_t} requires: $3~R~nnz(\T{X})$ flops to compute the MTTKRP using state-of-the-art libraries for sparse tensors~\cite{bader2007efficient}, $2~R^2~I_n$ flops to compute ${\M{A}^{(n)}}^T~\M{A}^{n}$ and $(d-1)~R^2$ flops to update $\M{C}^{(-n)}$ as in Equation~\eqref{eq:sustain_t_coeff}. As discussed in the matrix case, the dominant cost of Algorithm~\ref{alg:core} is $2~R^2~I_n$ flops. Overall, Algorithm~\ref{alg:sustain_t} requires: $3~d~R~nnz(\T{X}) + 4~R^2~\sum_{n=1}^d I_n
+ d~(d-1)~R^2$ flops. In our experiments, the first term, thus the computation of MTTKRP, dominates the total cost.

\subsection{Interpretation for phenotyping}
Given the EHRs of a certain cohort, we form a patient-by-diagnoses matrix $\M{X}$, whose $\M{X}(i,j)$ cell is the number of encounters of patient $i$ where encounter diagnosis $j$ was recorded. In that case, the patient membership vector $\M{U}(i, :)$ of $\mname_M$ provides the distinct levels of frequency of each one of the $R$ phenotypes throughout the medical history of the $i$-th patient. Likewise, each column $\M{V}(:, r)$ indicates the frequency levels of each medical feature w.r.t. the $r$-th phenotype. Table~\ref{table:anchor_pheno} summarizes a phenotype example that accounts for the largest share of heart failure patients. Finally, due to the integer box (i.e., $\{0, \dots, \tau\}$) constraints employed on the factor matrices, we can interpret the integer $\V{\lambda}(r)$ values as scaling up the input encounter counts for the $r$-th phenotype. Thus, phenotypes with higher $\V{\lambda}(r)$ values are expected to describe more persistent medical conditions, with higher number of associated encounters.

The above interpretation can be extended to the tensor case. Consider a tensor $\T{X}$ whose $\T{X}(i,j,k)$ cell defines the count of encounters of patient $i$ where medication $k$ was ordered for the patient with diagnosis $j$ as the order indication. Factorizing this tensor using $\mname_T$ yields a patient factor $\M{A}^{(1)}$ which can be interpreted similarly to the $\M{U}$ factor in the matrix case. Also, the factor matrices $\M{A}^{(2)}, \M{A}^{(3)}$ corresponding to diagnosis and medication or procedure phenotypes can be interpreted similarly to the $\M{V}$ factor in the matrix case. The same applies to the $\V{\lambda}(r)$ model values.

\section{Experiments}\label{sec:exp}
\subsection{Setup}
\subsubsection{Description of datasets}
Table~\ref{table:data_stats} summarizes statistics for the datasets used.

\noindent\textbf{Sutter:} This dataset corresponds to EHRs from Sutter Palo Alto Medical Foundation (PAMF) Clinics. The patients are 50 to 80 years old adults chosen for a heart failure study~\cite{choi2016using}. To form a patient-by-diagnosis matrix input, we extracted the number of encounter records with a specific diagnosis for each patient. To form a patient-diagnosis-medication tensor input, we used the medication orders, reflecting the ordered medications and the indicated diagnosis. We adopt standard medical concept groupers to group the available ICD-9 diagnosis codes~\cite{slee1978international} into Clinical Classification Software (CCS)~\cite{ccs} diagnostic categories (level 4). We also group the normalized drug names (i.e., combining all branded names and the generic name for a medication) based on unique therapeutic subclasses using the Anatomical Therapeutic Chemical Classification System.

\noindent\textbf{CMS:} We used a \textit{publicly-available} CMS Linkable 2008-2010 Medicare Data Entrepreneurs' Synthetic Public Use File (DE-SynPUF)~\footnote{These data can be downloaded from \url{https://www.cms.gov/Research-Statistics-Data-and-Systems/Downloadable-Public-Use-Files/SynPUFs/DE_Syn_PUF.html}} that 
contains three years of claim records synthesized (i.e., to protect privacy) from $5\%$ of the 2008 Medicare population. CMS creates twenty $5\%$ subsamples of the claims data. 
We used the carrier claims data available from DE-SynPUF for the patients belonging to Samples 1 \& 2. We increase the number of samples (i.e., number of patients) considered for the experiments related to assessing scalability. We used the diagnostic code information to build the input matrix and the diagnoses and procedures recorded to build the input tensor. In particular, we group the available ICD-9 diagnosis codes~\cite{slee1978international} into CCS~\cite{ccs} diagnostic categories (level 4) and use the CCS flat code grouper~\cite{ccs} to transform the CPT procedure codes available into procedure categories.

\begin{table}
\centering
\small
\begin{tabular}{c c c c}
\toprule 
\textbf{dataset} & \textbf{modes} & \textbf{size of modes} & \textbf{\#nnz ($\approx$Millions)} \\
\midrule
Sutter-matrix & Pat-Dx & $\num{259999} \times 576$ & $5.7$ \\
Sutter-tensor & Pat-Dx-Rx & $\num{248347} \times 552 \times 555$ & $5.4$ \\
CMS-matrix & Pat-Dx & $\num{197212} \times 583$ & $10.9$ \\
CMS-tensor & Pat-Dx-Proc & $\num{197143} \times 583 \times 239$ & $23.4$ \\
\bottomrule
\end{tabular}
\caption{\footnotesize For each dataset used, we list its name, nature of input modes, their sizes and the approximate number of non-zeros. Pat refers to patients, Dx to diagnoses, Rx to medications and Proc to procedures.}
\label{table:data_stats}
\end{table}

\subsubsection{Baselines}~\label{sec:exp_base}
Below, we describe our efforts to design competitive baseline methods producing the target models in Problems~\ref{eq:sustain_m} and~\ref{eq:sustain_t}, for the matrix and the tensor cases respectively.\\
\noindent\textbf{Round:} 
This baseline rounds the factor matrices from nonnegative matrix/tensor factorization. In the matrix case, we used the implementation of Nonnegative Matrix Factorization (NMF)~\cite{Kim2014-gw,kim2011fast} and projected 
all the entries of the resulting factor matrices to $\mathbb{Z}_{\tau}$. We also set $\V{\lambda}$ to an all-ones vector as NMF typically does not have the diagonal matrix $\M{\Lambda}$. A typical issue of naively rounding NMF solutions is that values that are lower than $0.5$ are rounded to $0$, so a potentially large part of model information can be lost. 

In the tensor case, we used the CP-ALS algorithm as in the Tensor Toolbox~\cite{TTB_Software}, adjusted to impose non-negativity constraints~\cite{kim2011fast} on the factor matrices. 
Also, in contrast to the NMF case, CP-ALS produces a $\V{\lambda}$ vector of nonnegative real values. 
In order to alleviate the effect of zeroing out values less than $0.5$ we compute the cube root of the $\V{\lambda}$ vector element-wise and form a vector $\V{\hat{\lambda}}$. Then, we absorb this scaling in the factor matrices by multiplying $\M{A}^{(n)}~ diag(\V{\hat{\lambda}}), \forall n = \{1, \dots, d\}$ where $d$ is the input tensor's order. Finally, we set $\V{\lambda}$ to an all-ones vector and project all the entries of the resulting factor matrices to $\mathbb{Z}_{\tau}$.

\noindent\textbf{Scale-and-round:} We design a more sophisticated scale-and-rounding heuristic which scales the factor matrices of the real-valued solutions before performing the rounding. This step further alleviates the problem of zeroing out values less than $0.5$.

In the matrix case, we define the scaling factor ${\gamma_2(j)}={\tau/max(\M{V}(:, j))}$. Then, ${\M{\tilde{V}}(:, j)}$ ${= round\left( \gamma_2(j)  \M{V}(:, j) \right)}$. 
Similarly, we define ${\gamma_1(j)}=\tau$ $/max{(\M{U}(:, j))}$. Then, ${\M{\tilde{U}}(:, j)}$ ${= round\left( \gamma_1(j)  \M{U}(:, j) \right)}$. Those steps scale-up the maximum value of each factor matrix column to reach the upper bound $\tau$.
Then, this excess scaling is absorbed into $\V{\lambda}$ as: ${\V{\lambda} = round(1/\gamma_1 \gamma_2)}$. 

In the tensor case, we absorb the scaling of the $\V{\lambda}$ output of the real-valued solution into the factor matrices as in \enquote{Round}, and extend the Scale-and-round matrix approach accordingly.

\noindent\textbf{AILS: Alternating Integer Least Squares approach:} 
We used the Integer Least Squares (ILS) with box constraints approach which is proposed in~\cite{Breen2012-ld,W_Chang2008-gt}. 
This approach was recently unified within an Integer Matrix Factorization framework~\cite{Dong2017-sy}. 
We exploit the redundancy among ILS problems targeting the same factor matrix, so that the QR factorization in the reduction phase is only computed once. Note that solving general ILS problems is NP-hard~\cite{Dong2017-sy}, which is reflected in the runtime of this method in the experiments.  
We enabled the extraction of the integer $\V{\lambda}$ values through an ILS by noticing that vectorizing the original problem as
{\small
\begin{equation*}
{\text{min}} \left\{ ||  vec(\M{U}~\M{\Lambda}~\M{V}^T) - vec(\M{X})||_F^2 \;\middle|\;
\M{\Lambda}\in \mathbb{Z}_{+}^{R\times R} \right\}
\end{equation*}}can be transformed to~\cite{brewer1978kronecker}: 
${\small {\text{min}} \left\{ ||  (\M{V} \odot \M{U}) \V{\lambda} - vec(\M{X})||_F^2 \;\middle|\;
\M{\lambda}\in \mathbb{Z}_{+}^{R} \right\}}$ which gives the ILS to solve for. Note that we attempted to extend this approach for tensor input; however, the materialization of the Khatri-Rao product of all the factor matrices failed due to out of memory problems even for the smallest target rank for both of the datasets used. To illustrate the magnitude of this issue, the size needed for the Khatri-Rao product of all factor matrices for Sutter data and $R=5$ is: $248347 * 552 * 555 * 5 * 8$ bytes $\approx 3$ Terabytes.


\subsubsection{Evaluation metrics}
We evaluate the methods under comparison in terms of the trade-off between execution time and accuracy for various target ranks considered ($R = \{5, 10, 20, 40\}$). Accuracy is measured in terms of fit: $1 - ||\M{X} - \M{\hat{X}}||_F^2 /
||\M{X}||_F^2$, where $\M{\hat{X}}$ is the re-constructed input through the model factors (this extends trivially to the tensor case); fit can be considered as the the proportion of data explained by the model.


\subsubsection{Initialization details} 
In all experiments, when we compare \mname and AILS, we provide them with the same initialization.

Regarding the accuracy-time trade-off evaluation, we initialize with several schemes and for each method we choose the one providing the highest fit. The schemes are the following: a) round heuristic, b) scale-and-round heuristic, c) random: random initialization with integers within the required range and $\V{\lambda}$ set to all-ones vector, d) random 
\& sampling: random initialization of the patients factor and sampling from the input data to populate the rest of the factors. In the matrix case, we initialize each $j$-th column of $\M{V}$ by random sampling of input patient vectors and scaling them to lie on $\mathbb{Z}_{\tau}$ if needed. 
In the tensor case, for each sampled slice $\T{X}(i, :, :)$, we populate each $j$-th component of $\M{A}^{(2)}, \M{A}^{(3)}$ by sampling the row and column of $\T{X}(i, :, :)$ with the maximum sum. Note that when we measure execution time for each approach, we do take into account the time spent for its initialization.

In the scalability evaluation, we initialize each method with 
the random \& sampling scheme (d) described above; this provided better starting points than using pure random initialization. For this experiment, we ignore the initialization time, since we want to focus on the methods' scalability behavior.


\subsubsection{Implementation details}
We used MatlabR2017b for our implementations, along with functionalities for sparse tensors from the \enquote{Tensor Toolbox}~\cite{TTB_Software} and for nonnegative matrix factorization from the \enquote{nonnegfac-matlab}~\cite{Kim2014-gw} toolbox. The ILS solver we use for the AILS baseline is included in the state-of-the-art MILES software~\cite{chang2007miles}.

The zero-lock problem 
refers to the case when a single column is zeroed out, thus zeroing out an entire rank-$1$ component of the solution. To avoid that in our scheme, 
we add the smallest perturbation possible ($+1$) to a randomly-chosen coordinate of the vector zeroed out. 


In both \mname and AILS, we break the iterations when the successive difference of the objective drops below $1e-4$. Finally, the parameter $\tau$ is set to $5$ driven by discussions with medical experts and similarity to many medical scoring systems.

\subsubsection{Hardware}
We conducted our experiments on a server running Ubuntu 14.04 with 1TB of RAM and four Intel E5-4620 v4 CPU's with a maximum clock frequency of 2.10GHz. Each of the processors contains $10$ cores with $2$ threads each.

\begin{figure}
\centering
\includegraphics[scale=0.3]{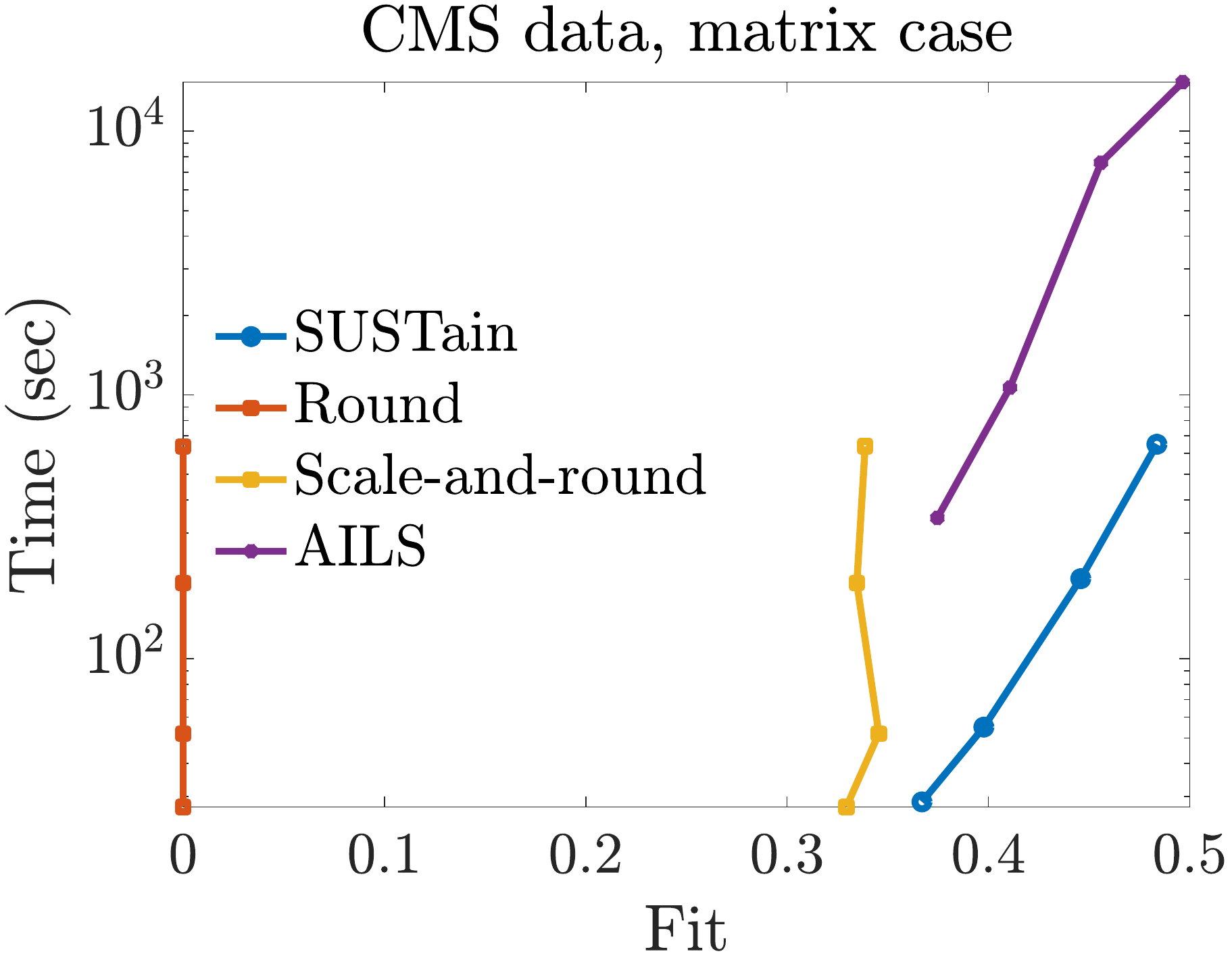}
\caption{\footnotesize Fit (range $[0, 1]$) vs time trade-off for varying target number of phenotypes $R=\{5, 10, 20, 40\}$ for the CMS matrix input. $\mname_M$ is at least an order of magnitude faster than the most accurate baseline (up to $38\times$ faster for $R=20$), while achieving the same level of accuracy. Also, $\mname_M$ achieves up to $14\%$ higher fit over scale-and-rounding heuristics.}
\label{fig:sustain_m_cms_tradeoff}
\end{figure}

\subsection{Matrix case experiments}
\noindent \textbf{Accuracy-Time trade-off:}
In Figure~\ref{fig:sustain_m_sutter_tradeoff}, we showcase the accuracy-time trade-off regarding the Sutter PAMF dataset. $\mname_M$ is at least $60\times$ faster ($R=40$) than the most accurate baseline (AILS). For $R=5$, $\mname_M$ achieves $425\times$ speedup over AILS: as compared to the \textbf{$\approx22$ minutes} spent by AILS, our approach executes in $\mathbf{\approx3}$ \textbf{seconds} for the same level of accuracy. Even for $R=\{10, 20\}$ $\mname_M$ achieves $98\times$ and $110\times$ faster computations than AILS. At the same time, $\mname_M$ achieves up to $16\%$ higher fit than the scale-and-round heuristic, operating on comparable running times. Note that for $R=5$, our approach is even faster (and more accurate) than the scale-and-round baseline as well, since initializing with random factors provided a better final fit than initializing with the scale-and-round result. We also remark that the naive round heuristic achieves a fit of zero, which is a by-product of zeroing out the majority of the model factor elements.

In Figure~\ref{fig:sustain_m_cms_tradeoff}, we provide the results of the same experiment regarding the CMS dataset. For the same level of accuracy, $\mname_M$ is at least an order of magnitude faster than AILS, and up to $38\times$ faster for $R=20$. It also achieves up to $14\%$ higher fit over the scale-and-rounding heuristic for comparable execution time.

\begin{table}
\centering
\small
\begin{tabular}{c c c c c}
\toprule 
\textbf{\#patients ($\approx$Thousands)} & $\mathbf{246}$ & $\mathbf{493}$ & $\mathbf{739}$ & $\mathbf{985}$ \\ 
\textbf{\#nnz ($\approx$Millions)} & $\mathbf{14}$ & $\mathbf{27}$ & $\mathbf{41}$ & $\mathbf{55}$ \\ \midrule
$\mname_M$ & $0.71$ & $0.95$ & $1.66$ & $2.82$ \\
Round / Scale-and-round & $4.4$ & $8.9$ & $12.9$ & $19.5$ \\
AILS & $339$ & $514$ & $940$ & $1254$ \\
\bottomrule
\end{tabular}
\caption{\footnotesize Running time (seconds) of one iteration for increasingly larger number of patients considered from the CMS data. Matrix case, $R=10$.}
\label{table:matrix_scale}
\end{table}
\noindent \textbf{Scaling for larger number of patients:} In Table~\ref{table:matrix_scale}, for fixed $R=10$, we measure a single iteration's time for increasing subsets of CMS patients. 
The NMF execution time is considered for Round and Scale-and-round heuristics, since their post-processing cost is negligible. $\mname_M$ can execute very fast (a single iteration in $\approx 3$ seconds) even for $\approx 985$ thousand patients. 

\begin{figure}
  \begin{minipage}[t]{0.49\linewidth}
    \centering
    \includegraphics[width=\linewidth]{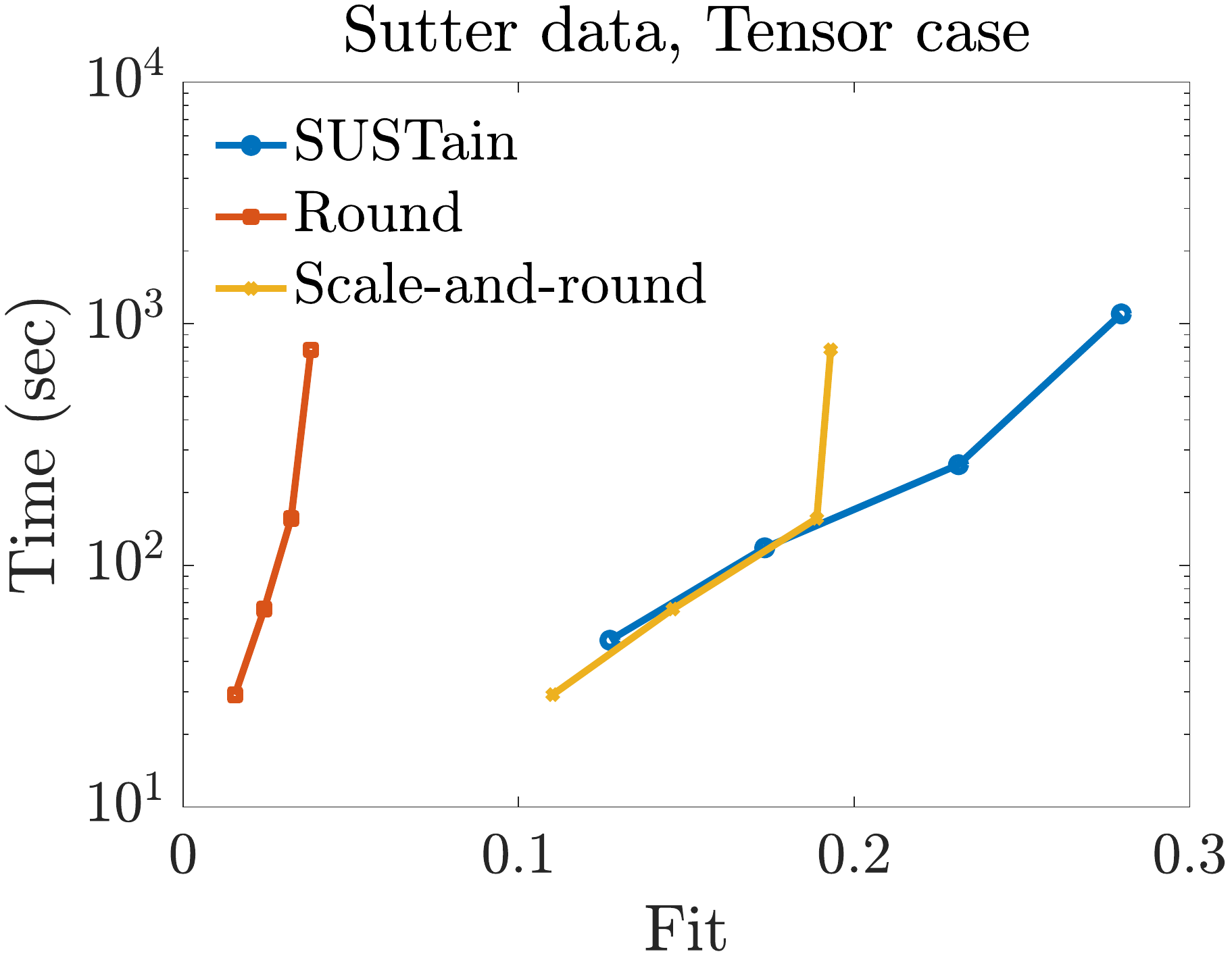}
  \end{minipage}
  \begin{minipage}[t]{0.49\linewidth}
    \centering
    \includegraphics[width=\linewidth]{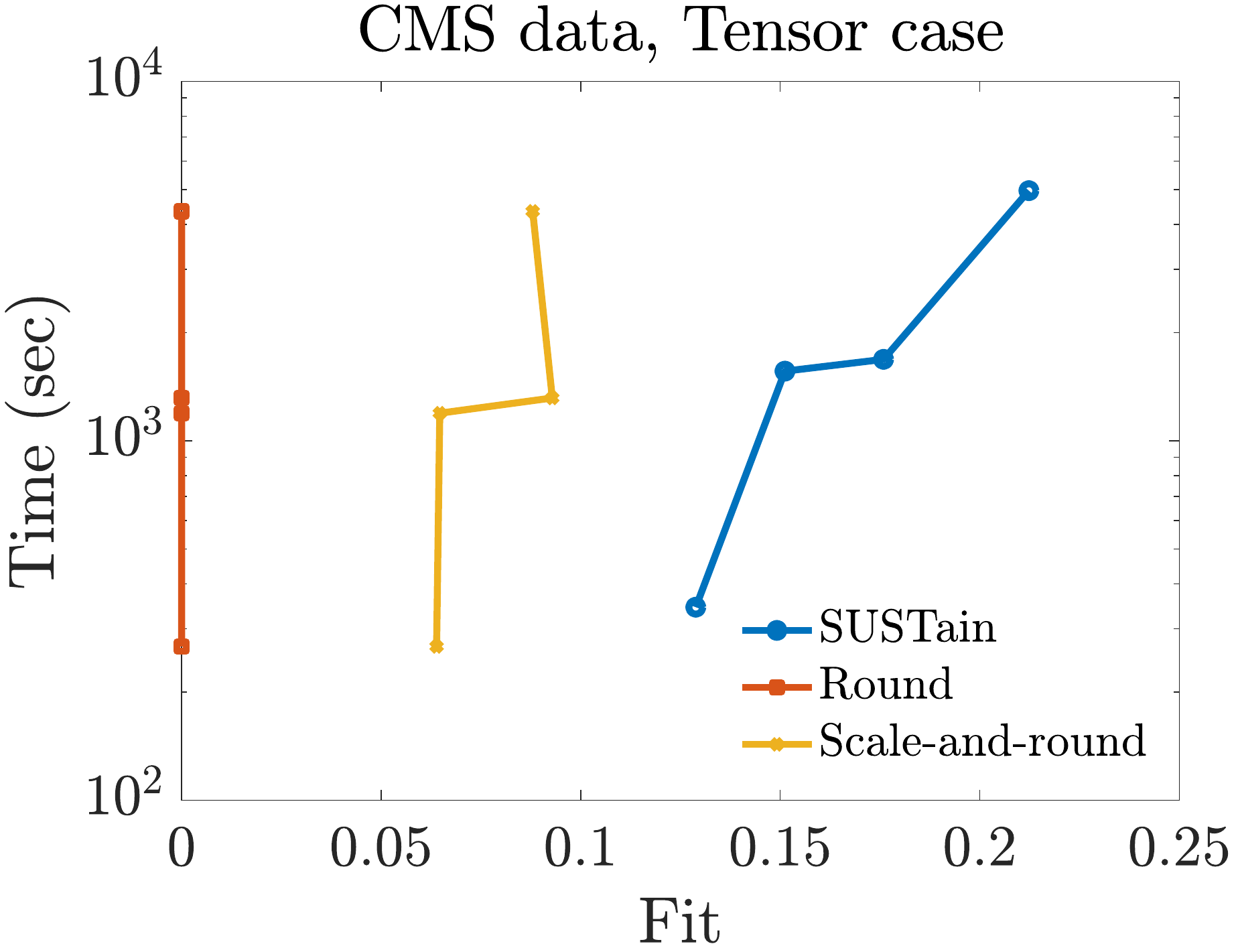}
  \end{minipage}
  \caption{\footnotesize Fit (range $[0, 1]$) vs time trade-off for varying target number of phenotypes $R=\{5, 10, 20, 40\}$ for the Sutter and the CMS tensor input. $\mname_T$  achieves up to $9\%$ and $12\%$ higher fit respectively over scale-and-rounding heuristics.}
  \label{fig:sustain_t_tradeoff}
\end{figure}
\subsection{Tensor case experiments}
\noindent \textbf{Accuracy-Time trade-off:} 
In Figure~\ref{fig:sustain_t_tradeoff}, we provide the fit-time trade-off for varying target rank of our input tensor datasets. As discussed in Section~\ref{sec:exp_base}, the extension of AILS approach to tensors cannot scale for any dataset or target rank considered. Overall, $\mname_T$ achieves up to $9\%$ and $12\%$ increase in fit over the scale-and-round heuristic w.r.t. the Sutter PAMF and CMS datasets respectively. Note that the fit of the scale-and-round approach decreases for successively increasing target rank values (e.g., transitioning from $R=20$ to $R=40$ for CMS data). 
This indicates that heuristic approaches which simply post-process real-valued solutions may not fully exploit the available target rank.

\begin{table}
\centering
\small
\begin{tabular}{c c c c c}
\toprule 
\textbf{\#patients ($\approx$Thousands)} & $\mathbf{246}$ & $\mathbf{493}$ & $\mathbf{739}$ & $\mathbf{985}$ \\ 
\textbf{\#nnz ($\approx$Millions)} & $\mathbf{29}$ & $\mathbf{58}$ & $\mathbf{88}$ & $\mathbf{117}$ \\ \midrule
$\mname_T$ & $38.5$ & $76.9$ & $115$ & $151$ \\
Round / Scale-and-round & $39.6$ & $78$ & $117$ & $157$ \\
\bottomrule
\end{tabular}
\caption{\footnotesize Running time (seconds) of one iteration for increasingly larger number of patients considered from the CMS data. Tensor case, $R=10$.}
\label{table:tensor_scale}
\end{table}
\noindent \textbf{Scaling for larger number of patients:}
In Table~\ref{table:tensor_scale}, we report the time spent for one iteration of increasingly larger subset of patients considered from the CMS data, with fixed target rank ($R=10$). The time measured for the heuristic approaches corresponds to the execution time of CP-ALS, since the post-processing cost is negligible. We observe that $\mname_T$ achieves linear scale-up w.r.t. increasing number of patients. We also remark that the dominant cost in both $\mname_T$ and the CP-ALS is the MTTKRP computation, which explains the comparable running time. 

\subsection{Case study on Phenotyping HF patients}
Cardiovascular disease (CVD) is the leading cause of death worldwide 
and heart failure (HF) is a dominant cause of morbidity and mortality. HF is traditionally characterized by reduced ejection fraction (HFrEF) and preserved ejection fraction (HFpEF). But, recent evidence suggests that HF is more heterogeneous than is reflected by ejection fraction. We used SUSTain to explore this heterogeneity in an incident HF cohort.

\noindent\textbf{Cohort and data selection:}
We select only the HF case patients 
from the Sutter PAMF dataset. For each incident HF case, we extracted data in the 12-months before and the 12-months after the initial HF diagnosis date, which resulted in $70,531$ clinical encounters. 
We used all the data modalities available, i.e., medication orders and indications and encounter diagnoses. 
The size of the resulting (patient-by-diagnosis-by-medication) tensor is $3,497 \times 396 \times 367$; the tensor contains a total of $92,662$ non-zero elements.

\noindent\textbf{Choosing the number of phenotypes:} 
We use the stability-driven criterion introduced in~\cite{wu2016stability}. The intuition behind this criterion is in promoting a target rank 
for which several runs with different initial points return reproducible factors. We choose the diagnosis factor matrix as the factor under assessment. Let $\M{D}_1$ and $\M{D}_2$ be the diagnosis factor matrix for $2$ different runs with the same target rank. Then, the cross-correlation matrix $\M{C}\in\mathbb{R}^{R\times R}$ is computed between the columns of $\M{D}_1, \M{D}_2$ and the dissimilarity between them is computed as~\cite{wu2016stability}:
{\small
\[ diss(\M{D}_1, \M{D}_2) = \frac{1}{2 R} \left( 2R - \sum_{j=1}^R max_{1\leq k \leq R} \M{C}(k,j) - \sum_{k=1}^R max_{1\leq j \leq R} \M{C}(k,j) \right)  \]
}Note that when $\M{D}_1$ can be transformed to $\M{D}_2$ by column permutation, then $diss(\M{D}_1, \M{D}_2) = 0$. If $B$ is the number of repetitions for each target rank, then the following relation computes the average dissimilarity over $B (B-1) / 2$ pairs of resulting factors:
{\small
\[ Y(R) = \frac{2}{B (B-1)} \sum_{1\leq b < b' \leq B} diss(\M{D}_b, \M{D}_{b'}) \]
}We used the \enquote{staNMF} toolbox 
to compute the above score for each target rank on the range $\{5, \dots, 20\}$. 
The input to $\mname_T$ were $B=20$ initial points of the round heuristic. 
$R=15$ phenotypes were selected based on the above criterion. For the target rank chosen, we pick the solution yielding the highest fit.


\begin{table}
\centering
\small
\begin{tabular}{c c c c c}
\toprule 
\textbf{method} & $\mathbf{\#nnz(\M{A}^{(1)})}$ & $\mathbf{\#nnz(\M{A}^{(2)})}$ & $\mathbf{\#nnz(\M{A}^{(3)})}$ & \textbf{fit}   \\
\midrule
$\mname_T$ & $3,438$ & $54$ & $88$ & $0.261$ \\
NN CP-ALS & $3,497$ & $60$ & $90$ & $0.175$\\
\bottomrule
\end{tabular}
\caption{\footnotesize $\mname_T$ achieves $\approx 8.6\%$ increase in fit than a Nonnegative CP-ALS model truncated to achieve the same level of sparsity. The result refers to the HF case study for $R=15$.}
\label{table:hf_tensor_density}
\end{table}

\noindent\textbf{\mname provides concise and accurate solutions:}
We observed that besides preserving the input data properties and providing a natural interpretation for medical experts, \mname implicitly imposes sparse factors. To assess the factors' conciseness, we compare their fit with the achieved fit of the real-valued model (NN CP-ALS), which is post-processed to achieve factor sparsity (as would be done by a practitioner). For each of the feature factors (diagnosis and medication) of the real-valued model, we only consider the top-$k$ elements for each column (i.e., most important elements of each phenotype). For the patient factor, we consider the top-$k$ elements for each row (i.e., most important phenotypes for each patient). In each case, the value of $k$ is chosen so that the sparsity level is close to the one achieved by $\mname_T$. We provide the results in Table~\ref{table:hf_tensor_density}, where we notice that for the same level of sparsity, $\mname_T$ achieves $\approx 8.6\%$ increase in fit. Thus, the integer factors of $\mname_T$ decompose the input more accurately for the same level of sparsity than the real-valued counterpart.

\begin{table}
\centering
\footnotesize
\begin{tabular}{c c}
\toprule 
\textbf{HF with reduced LVEF (HFrEF)}
& \textbf{Score} \\ \midrule 
Rx\_Loop Diuretics & $3$ \\
Dx\_Congestive heart failure & $1$ \\
Rx\_ACE Inhibitors & $1$ \\
Rx\_Alpha-Beta Blockers & $1$\\
Rx\_Potassium & $1$ \\
\midrule
\textbf{Hypertension}
& \textbf{Score} \\ \midrule 
Rx\_ACE Inhibitors & $3$ \\
Dx\_Essential hypertension & $1$ \\
Rx\_Alpha-Beta Blockers & $1$ \\
Rx\_Beta Blockers Cardio-Selective & $1$ \\
Rx\_Calcium Channel Blockers & $1$ \\
Rx\_HMG CoA Reductase Inhibitors & $1$ \\
Rx\_Loop Diuretics & $1$ \\
Rx\_Thiazides and Thiazide-Like Diuretics & $1$ \\ 
\midrule 
\textbf{Hypertension (more difficult to control)}
& \textbf{Score} \\ \midrule 
Rx\_Angiotensin II Receptor Antagonists & $2$ \\
Rx\_Beta Blockers Cardio-Selective & $2$ \\
Rx\_Calcium Channel Blockers & $2$\\
Dx\_Essential hypertension & $1$ \\
Rx\_Antiadrenergic Antihypertensives & $1$\\
Rx\_Loop Diuretics & $1$ \\
Rx\_Potassium & $1$ \\
\bottomrule
\end{tabular}
\caption{\footnotesize 
Representative phenotypes extracted by $\mname_T$ for our HF case study. The score of each feature indicates its relative frequency within the phenotype.
The prefix for each feature indicates whether it corresponds to a medication (Rx) or a diagnosis (Dx). 
A cardiologist provided phenotype annotations and validated that: the top-most phenotype is aligned to guideline-based management of HF with reduced LVEF (HFrEF), the next one corresponds to typical hypertensive patients (common risk factor of HF) and the last one corresponds to hypertensive patients being more difficult to control.}
\label{table:pheno2}
\end{table}

\noindent\textbf{Phenotype discovery:} In Table~\ref{table:anchor_pheno} and Table~\ref{table:pheno2}, we provide representative phenotypes extracted through our method. 
A subset of annotations provided by the cardiologist are as follows: hyperlipidemia (the one in Table~\ref{table:anchor_pheno}), HF with reduced LVEF (HFrEF), hypertension (HTN), HTN which is more difficult to control, persistent and chronic atrial fibrillation, depression, diabetes, comorbidities of aging, prior pulmonary embolism. Overall, 13 out of 15 phenotype candidates were annotated as clinically meaningful phenotypes related to heart failure.

\section{Related Work}
\noindent {\bf Discrete factorization-based approaches:} Dong et al.~\cite{Dong2017-sy} proposed an Integer Matrix Factorization framework via solving Integer Least Squares subproblems. As we experimentally evaluated, this approach is orders of magnitude slower than \mname while achieving the same level of accuracy. 
Kolda and O'Leary~\cite{Kolda1998-up} proposed a Semidiscrete Matrix Decomposition into factors containing ternary values ($\{-1, 0, 1\}$). Despite its demonstrated success for compression purposes, a direct application of this approach would introduce negative values into the factors, thus hurting interpretability for nonnegative input. Finally, several prior works target binary factorization 
(e.g.,~\cite{koyuturk2006nonorthogonal,Zhang2007-ex,Shen2009-ls,Miettinen2008-qj,miettinen2011boolean,lian2017discrete}). In contrast to strictly binary factors, \mname captures the quantity embedded in the input data, which reveals important information (e.g., relative phenotype prevalence and associated feature frequencies).

\noindent {\bf Unsupervised Phenotyping:} Extensive prior work applies factorization techniques for unsupervised phenotyping
(e.g.,~\cite{Ho2014-ml,Ho2014-mh,Wang2015-pa,Perros2015-te,Joshi2016-kt,Perros2017-dh}). However, no work  considered extracting scoring-based phenotypes to facilitate their interpretation by domain experts.

\noindent {\bf HALS fitting algorithms:} Our fitting algorithms follow the intuition of Hierarchical Alternating Least Squares (HALS) framework~\cite{Cichocki2009-jg} (aka rank-one residue iteration~\cite{Ho_undated-wp}), which enables formulating the solution for each $k$-th rank-$1$ component separately. However, plain HALS does not tackle the challenges involved with either imposing integer constraints or solving for the vector $\V{\lambda}$.

\section{Conclusions}\label{sec:concl}

The accuracy and scalability of \mname on \enquote{native} integer data derives from two key insights.
One is expected: just rounding or applying related transformations to real-valued solutions is inherently limited.
The second may be more surprising: while discrete constraints might appear to make the problem more challenging, in fact, a careful organization of the problem into subparts can mitigate that complexity.
In our case, we identify a problem partitioning of integer-constrained subproblems that leads to an optimal and efficient solution;
and, we also define the order of alternating updates so as to enable reuse of shared intermediate results.
Consequently, \mname outperforms several baselines on both synthetic (publicly-available) and real EHR data, showing either a better fit or orders-of-magnitude speedups at a comparable fit.

Moving forward, there are many other sources of integer values in real-world data.
These include, for instance, ordinal values.
Thus, whereas this paper targets event counts, extensions for other cases is a ripe target for future work.



%

\bibliographystyle{abbrv}
\bibliography{sigproc}

\begin{thebibliography}{10}

\bibitem{ccs}
Clinical classifications software (ccs) for icd-9-cm.
\newblock \url{https://www.hcup-us.ahrq.gov/toolssoftware/ccs/ccs.jsp}, 2017.
\newblock Accessed: 2017-02-11.

\bibitem{bader2007efficient}
B.~W. Bader and T.~G. Kolda.
\newblock Efficient matlab computations with sparse and factored tensors.
\newblock {\em SIAM Journal on Scientific Computing}, 30(1):205--231, 2007.

\bibitem{TTB_Software}
B.~W. Bader, T.~G. Kolda, et~al.
\newblock Matlab tensor toolbox version 2.6.
\newblock Available online, February 2015.

\bibitem{boyd2004convex}
S.~Boyd and L.~Vandenberghe.
\newblock {\em Convex optimization}.
\newblock Cambridge university press, 2004.

\bibitem{Breen2012-ld}
S.~Breen and X.-W. Chang.
\newblock Column reordering for {Box-Constrained} integer least squares
  problems.
\newblock Apr. 2012.

\bibitem{brewer1978kronecker}
J.~Brewer.
\newblock Kronecker products and matrix calculus in system theory.
\newblock {\em IEEE Transactions on circuits and systems}, 25(9):772--781,
  1978.

\bibitem{bro1998least}
R.~Bro and N.~D. Sidiropoulos.
\newblock Least squares algorithms under unimodality and non-negativity
  constraints.
\newblock {\em Journal of Chemometrics}, 12(4):223--247, 1998.

\bibitem{carroll1970analysis}
J.~D. Carroll and J.-J. Chang.
\newblock Analysis of individual differences in multidimensional scaling via an
  n-way generalization of "eckart-young" decomposition.
\newblock {\em Psychometrika}, 35(3):283--319, 1970.

\bibitem{chang2007miles}
X.-W. Chang and T.~Zhou.
\newblock Miles: Matlab package for solving mixed integer least squares
  problems.
\newblock {\em GPS Solutions}, 11(4):289--294, 2007.
\newblock Last updated: June 2016.

\bibitem{choi2016using}
E.~Choi, A.~Schuetz, W.~F. Stewart, and J.~Sun.
\newblock Using recurrent neural network models for early detection of heart
  failure onset.
\newblock {\em Journal of the American Medical Informatics Association},
  24(2):361--370, 2016.

\bibitem{Cichocki2009-jg}
A.~Cichocki and A.-H. Phan.
\newblock Fast local algorithms for large scale nonnegative matrix and tensor
  factorizations.
\newblock {\em IEICE transactions on fundamentals of electronics,
  communications and computer sciences}, 92(3):708--721, 2009.

\bibitem{denny2010phewas}
J.~C. Denny, M.~D. Ritchie, M.~A. Basford, J.~M. Pulley, L.~Bastarache,
  K.~Brown-Gentry, D.~Wang, D.~R. Masys, D.~M. Roden, and D.~C. Crawford.
\newblock Phewas: demonstrating the feasibility of a phenome-wide scan to
  discover gene--disease associations.
\newblock {\em Bioinformatics}, 26(9):1205--1210, 2010.

\bibitem{Dong2017-sy}
B.~Dong, M.~M. Lin, and H.~Park.
\newblock Integer matrix approximation and data mining.
\newblock {\em Journal of scientific computing}, pages 1--27, Sept. 2017.

\bibitem{gillis2011nonnegative}
N.~Gillis et~al.
\newblock Nonnegative matrix factorization: Complexity, algorithms and
  applications.
\newblock {\em Unpublished doctoral dissertation, Universit{\'e} catholique de
  Louvain. Louvain-La-Neuve: CORE}, 2011.

\bibitem{golub2013matrix}
G.~H. Golub and C.~F. Van~Loan.
\newblock {\em Matrix Computations}, volume~3.
\newblock JHU Press, 2013.

\bibitem{harshman1970foundations}
R.~A. Harshman.
\newblock Foundations of the parafac procedure: Models and conditions for an
  "explanatory" multi-modal factor analysis.
\newblock 1970.

\bibitem{Ho2014-ml}
J.~C. Ho, J.~Ghosh, S.~R. Steinhubl, W.~F. Stewart, J.~C. Denny, B.~A. Malin,
  and J.~Sun.
\newblock Limestone: high-throughput candidate phenotype generation via tensor
  factorization.
\newblock {\em Journal of biomedical informatics}, 52:199--211, Dec. 2014.

\bibitem{Ho2014-mh}
J.~C. Ho, J.~Ghosh, and J.~Sun.
\newblock Marble: High-throughput phenotyping from electronic health records
  via sparse nonnegative tensor factorization.
\newblock In {\em Proceedings of the 20th {ACM} {SIGKDD} International
  Conference on Knowledge Discovery and Data Mining}, KDD '14, pages 115--124,
  New York, NY, USA, 2014. ACM.

\bibitem{Ho_undated-wp}
N.-D. Ho.
\newblock {\em Nonnegative matrix factorization algorithms and applications}.
\newblock PhD thesis, PhD thesis, Universit{\'e} catholique de Louvain, 2008.

\bibitem{Joshi2016-kt}
S.~Joshi, S.~Gunasekar, D.~Sontag, and G.~Joydeep.
\newblock Identifiable phenotyping using constrained {Non-Negative} matrix
  factorization.
\newblock In {\em Machine Learning for Healthcare Conference}, pages 17--41,
  Dec. 2016.

\bibitem{Kim2014-gw}
J.~Kim, Y.~He, and H.~Park.
\newblock Algorithms for nonnegative matrix and tensor factorizations: A
  unified view based on block coordinate descent framework.
\newblock {\em Journal of Global Optimization}, 58(2):285--319, Feb. 2014.

\bibitem{kim2011fast}
J.~Kim and H.~Park.
\newblock Fast nonnegative matrix factorization: An active-set-like method and
  comparisons.
\newblock {\em SIAM Journal on Scientific Computing}, 33(6):3261--3281, 2011.

\bibitem{kolda2009tensor}
T.~G. Kolda and B.~W. Bader.
\newblock Tensor decompositions and applications.
\newblock {\em SIAM review}, 51(3):455--500, 2009.

\bibitem{Kolda1998-up}
T.~G. Kolda and D.~P. O'Leary.
\newblock A semidiscrete matrix decomposition for latent semantic indexing
  information retrieval.
\newblock {\em ACM Transactions on Information and System Security},
  16(4):322--346, Oct. 1998.

\bibitem{kolda2008scalable}
T.~G. Kolda and J.~Sun.
\newblock Scalable tensor decompositions for multi-aspect data mining.
\newblock In {\em Data Mining, 2008. ICDM'08. Eighth IEEE International
  Conference on}, pages 363--372. IEEE, 2008.

\bibitem{koyuturk2006nonorthogonal}
M.~Koyut{\"u}rk, A.~Grama, and N.~Ramakrishnan.
\newblock Nonorthogonal decomposition of binary matrices for bounded-error data
  compression and analysis.
\newblock {\em ACM Transactions on Mathematical Software (TOMS)}, 32(1):33--69,
  2006.

\bibitem{Lee1999-kk}
D.~D. Lee and H.~S. Seung.
\newblock Learning the parts of objects by non-negative matrix factorization.
\newblock {\em Nature}, 401(6755):788--791, Oct. 1999.

\bibitem{lian2017discrete}
D.~Lian, R.~Liu, Y.~Ge, K.~Zheng, X.~Xie, and L.~Cao.
\newblock Discrete content-aware matrix factorization.
\newblock In {\em Proceedings of the 23rd ACM SIGKDD International Conference
  on Knowledge Discovery and Data Mining}, pages 325--334. ACM, 2017.

\bibitem{miettinen2011boolean}
P.~Miettinen.
\newblock Boolean tensor factorizations.
\newblock In {\em Data Mining (ICDM), 2011 IEEE 11th International Conference
  on}, pages 447--456. IEEE, 2011.

\bibitem{Miettinen2008-qj}
P.~Miettinen, T.~Mielik{\"a}inen, A.~Gionis, G.~Das, and H.~Mannila.
\newblock The discrete basis problem.
\newblock {\em IEEE transactions on knowledge and data engineering},
  20(10):1348--1362, Oct. 2008.

\bibitem{Perros2015-te}
I.~Perros, R.~Chen, R.~Vuduc, and J.~Sun.
\newblock Sparse hierarchical tucker factorization and its application to
  healthcare.
\newblock In {\em Data Mining (ICDM), 2015 IEEE International Conference on},
  pages 943--948. IEEE, 2015.

\bibitem{Perros2017-dh}
I.~Perros, E.~E. Papalexakis, F.~Wang, R.~Vuduc, E.~Searles, M.~Thompson, and
  J.~Sun.
\newblock {SPARTan}: Scalable {PARAFAC2} for large \& sparse data.
\newblock In {\em Proceedings of the 23rd {ACM} {SIGKDD} International
  Conference on Knowledge Discovery and Data Mining}, KDD '17, pages 375--384.
  ACM, 2017.

\bibitem{Richesson2016-ar}
R.~L. Richesson, J.~Sun, J.~Pathak, A.~N. Kho, and J.~C. Denny.
\newblock Clinical phenotyping in selected national networks: demonstrating the
  need for high-throughput, portable, and computational methods.
\newblock {\em Artificial intelligence in medicine}, 71:57--61, July 2016.

\bibitem{Shen2009-ls}
B.-H. Shen, S.~Ji, and J.~Ye.
\newblock Mining discrete patterns via binary matrix factorization.
\newblock In {\em Proceedings of the 15th {ACM} {SIGKDD} International
  Conference on Knowledge Discovery and Data Mining}, KDD '09, pages 757--766,
  New York, NY, USA, 2009. ACM.

\bibitem{slee1978international}
V.~N. Slee.
\newblock The international classification of diseases: ninth revision (icd-9).
\newblock {\em Annals of internal medicine}, 88(3):424--426, 1978.

\bibitem{takapoui2017simple}
R.~Takapoui, N.~Moehle, S.~Boyd, and A.~Bemporad.
\newblock A simple effective heuristic for embedded mixed-integer quadratic
  programming.
\newblock {\em International Journal of Control}, pages 1--11, 2017.

\bibitem{Ustun2017-ri}
B.~Ustun and C.~Rudin.
\newblock Optimized risk scores.
\newblock In {\em Proceedings of the 23rd {ACM} {SIGKDD} International
  Conference on Knowledge Discovery and Data Mining}, pages 1125--1134. ACM,
  13~Aug. 2017.

\bibitem{W_Chang2008-gt}
X.~w.~Chang and Q.~Han.
\newblock Solving {Box-Constrained} integer least squares problems.
\newblock {\em IEEE Transactions on Wireless Communications}, 7(1):277--287,
  Jan. 2008.

\bibitem{Wang2015-pa}
Y.~Wang, R.~Chen, J.~Ghosh, J.~C. Denny, A.~Kho, Y.~Chen, B.~A. Malin, and
  J.~Sun.
\newblock Rubik: Knowledge guided tensor factorization and completion for
  health data analytics.
\newblock In {\em Proceedings of the 21th {ACM} {SIGKDD} International
  Conference on Knowledge Discovery and Data Mining}, KDD '15, pages
  1265--1274, New York, NY, USA, 2015. ACM.

\bibitem{wu2016stability}
S.~Wu, A.~Joseph, A.~S. Hammonds, S.~E. Celniker, B.~Yu, and E.~Frise.
\newblock Stability-driven nonnegative matrix factorization to interpret
  spatial gene expression and build local gene networks.
\newblock {\em Proceedings of the National Academy of Sciences},
  113(16):4290--4295, 2016.

\bibitem{Zhang2007-ex}
Z.~Zhang, T.~Li, C.~Ding, and X.~Zhang.
\newblock Binary matrix factorization with applications.
\newblock In {\em Seventh {IEEE} International Conference on Data Mining
  ({ICDM} 2007)}, pages 391--400, Oct. 2007.

\end{thebibliography}

\end{document}